\documentclass{bmvc2k}

\usepackage{graphics}
\usepackage{amsfonts}
\usepackage{booktabs}
\usepackage[table,dvipsnames]{xcolor}

\DeclareMathOperator*{\argmin}{\arg\!\min}

\title{Structured Prediction of 3D Human Pose with Deep Neural Networks}

\addauthor{Bugra Tekin$^{*}$}{bugra.tekin@epfl.ch}{1}
\addauthor{Isinsu Katircioglu$^{*}$}{isinsu.katircioglu@epfl.ch}{1}
\addauthor{Mathieu Salzmann}{mathieu.salzmann@epfl.ch}{1}
\addauthor{Vincent Lepetit}{lepetit@icg.tugraz.at}{2}
\addauthor{Pascal Fua}{pascal.fua@epfl.ch}{1}

\addinstitution{
 CVLab\\
 EPFL,\\
 Lausanne, Switzerland
}
\addinstitution{
 CVARLab\\
 TU Graz,\\
 Graz, Austria
}

\runninghead{Tekin et al.}{Structured Prediction of 3D Human Pose}

\graphicspath{{images/}}

\newcommand{\comment}[1]{}

\newcommand{\btrmk}[1]{\textcolor{black}{\bf bt: #1}}
\newcommand{\bt}[1]{{\color{black} #1}}

\newcommand{\pfrmk}[1]{\textcolor{black}{\bf pf: #1}}
\newcommand{\pf}[1]{{\color{black} #1}}

\newcommand{\ms}[1]{{\color{black} #1}}

\newcommand{\vincent}[1]{{\color{black} #1}}

\begin{document}

\maketitle

\begin{abstract}
  
Most recent  approaches to monocular 3D  pose estimation rely on  Deep Learning.
They either train a Convolutional Neural  Network to directly regress from image
to 3D pose,  which ignores the dependencies between human  joints, or 
model these dependencies  via a max-margin structured  learning framework, which
involves a high computational cost at inference time.

In  this  paper,  we  introduce  a Deep  Learning  regression  architecture  for
structured prediction of  3D human pose from monocular images  that relies on an
overcomplete auto-encoder to learn  a high-dimensional latent pose representation
and  account for  joint dependencies.   We demonstrate  that our  approach
outperforms state-of-the-art  ones both in  terms of structure  preservation and
prediction accuracy.

\end{abstract}


\section{Introduction} \label{sec:intro}

3D human pose can now be  estimated reliably by training algorithms to
exploit     depth    data~\cite{Girshick11,Shotton11}     or    video
sequences~\cite{Andriluka10,Huang14,Tekin16}.   However,  estimating \vincent{such a} 3D  pose
from single  ordinary images remains  challenging because of  the many
ambiguities inherent to monocular  3D reconstruction\vincent{,
  including occlusions, complex backgrounds, and, more generally, the loss of
depth information resulting from the projection from 3D to 2D.}


\ms{These ambiguities can be mitigated by  exploiting the structure of the human
  pose,   that  is,   the  dependencies   between  the   different  body   joint
  locations.  This has  been done  by explicitly  enforcing physical constraints  at test
  time~\cite{Sminchisescu03a,Salzmann10a}  and by  data-driven  priors over  the
  pose space~\cite{Sidenbladh00,Urtasun05b,Gall10}.  Recently, dependencies have
  been   modeled  within   a  Deep   Learning  framework   using  a   max-margin
  formalism~\cite{Li15a},   which   resulted  in   state-of-the-art   prediction
  accuracy.  While effective,  these  methods  suffer from  the  fact that  they
  require solving  a computationally expensive optimization  problem to estimate
  the 3D pose.

By contrast,  regression-based methods,
such  as~\cite{Li14a}, directly  \vincent{and efficiently}  predict the  3D pose
given the  input image.   While this  often comes  at the  cost of  ignoring the
underlying  structure,  several  methods  have  been  proposed  to  account  for
it~\cite{Salzmann10c,Ionescu14a}.   In~\cite{Ionescu14a}, this  was achieved  by
making use of Kernel Dependency Estimation~(KDE)~\cite{Cortes05,Weston02}, which
maps  both input  and output  to high-dimensional  Hilbert spaces  and learns  a
mapping  between these  spaces.   Because this  approach  relies on  handcrafted
features  and  does  not  exploit  the  power  of  Deep  Learning,  it  somewhat
under-performs more recent CNN-based techniques~\cite{Li14a,Li15a}.}



\begin{figure}[t] 
	\centering 
	\includegraphics[width=1.0\columnwidth]{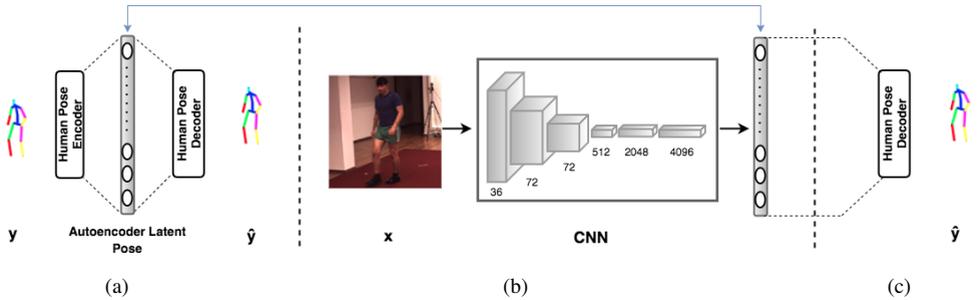} \\
	\footnotesize \hspace{0.55cm} (a) \hspace{4.8cm} (b) \hspace{4.6cm} (c)
	\caption{Our architecture for the structured prediction of the 3D human pose. 
		{\bf (a)} An auto-encoder whose hidden layers have  a larger dimension 
		than both its input and  output layers is pretrained. In practice  we use either this  
		one or more sophisticated  versions   that  are   described  in  more   
		detail  in Section~\ref{sec:encoder} {\bf (b)} A CNN is mapped into  
		the  latent representation learned by the auto-encoder.   
		{\bf  (c)} the latent representation is mapped back to the original 
		pose space using the decoder.}
	
	\label{fig:hpe}
\end{figure}

In  this paper,  we  demonstrate that  we  can account  for  \ms{the human  pose
  structure} within a deep learning  framework by first training an overcomplete
auto-encoder  that projects  body joint  positions to  a high  dimensional space
represented by its middle layer,  as depicted by Fig.~\ref{fig:hpe}(a).  We then
learn a  CNN-based mapping from  the input  image to this  high-dimensional pose
representation as shown in Fig.~\ref{fig:hpe}(b). \ms{This is inspired by KDE in
  that} it can be understood as  replacing kernels by the auto-encoder layers to
predict the pose parameters in a high dimensional space \bt{that encodes complex
  dependencies between different body parts}.  As a result, it enforces implicit
constraints on the human pose, preserves the human body statistics, and improves
prediction accuracy, as  will be demonstrated by  our experiments.  \ms{Finally,
  as illustrated in Fig.~\ref{fig:hpe}(c), we connect the decoding layers of the
  auto-encoder  to  this  network,  and  fine-tune  the  whole  model  for  pose
  estimation.}
  
  \comment{ Furthermore,  the  latent
  representation  our encoder  learns is  sparse and  its decoder  only performs
  linear operations.  Thus, unlike other  structured learning approaches such as
  maximum-margin methods~\cite{Li15a} that require an exhaustive search over the
  pose space, ours does does not impose a substantial computational burden.}

\comment{In  this paper,  we  demonstrate  that we  can  account for  these
dependencies  in  a  deep  learning framework  by  first  training  an
overcomplete auto-encoder that projects body  joint positions to a high
dimensional but sparse  pose space. As in the same  spirit with kernel
dependency estimation using the  ``kernel trick'' on output variables,
we predict  the parameters of the  pose in the high  dimensional space
instead  of directly  predicting the  3D joint  locations. We  finally
recover  the  3D  pose of  the  person  by  the  decoder part  of  our
auto-encoder.  As   this  scheme,  shown   in  Fig.~\ref{fig:overview},
enforces  implicit constraints  on the  human pose,  it preserves  the
human body statistics and improves the accuracy of the predictions, as
will  be  demonstrated by  our  experiments.   Furthermore, since  the
latent representation  our encoder  learns is  sparse and  its decoder
only  performs linear  operations, it  does not  impose a  substantial
computational burden unlike other  structured learning approaches such
as  maximum-margin  methods~\cite{Li15a}  which   requires  to  do  an
exhaustive search over  the space of training  poses.}

\comment{  In this  paper, we  demonstrate  we can  account for  these
dependencies  in  a  deep  learning framework  by  first  training  an
overcomplete auto-encoder that projects body  joint positions to a high
dimensional but  sparse space and  extracts the latent  pose embedding
layer.\pfrmk{Latent  pose  embedding layer  is  not  obvious. Can  you
clarify? How  about a figure?}  We  then integrate it into  a complete
CNN-based regressor that directly regresses from image to 3D pose.  We
will  show  that  it   outperforms  the  above-mentioned  methods  and
preserves  human   body  statistics  Furthermore,  since   the  latent
representation our encoder learns is  very sparse and its decoder only
performs  linear   operations,  it  does  not   impose  a  substantial
computational burden.\pfrmk{Unlike?}}

In  short, our  contribution  is to  show that  combining  traditional CNNs  for
supervised  learning with  auto-encoders  for structured  learning preserves  the
power of  CNNs while  also accounting for  dependencies, resulting  in increased
performance.  In the remainder of the paper, we first briefly discuss earlier 
approaches. We then present our structured prediction approach in more detail and 
finally demonstrate that it outperforms state-of-the-art methods on the Human3.6m dataset.



\section{Related Work}

\ms{Following recent  trends in  Computer Vision, human  pose estimation  is now
  usually formulated  within a  Deep Learning framework.   The switch  away from
  earlier  representations  started  with  2D  pose  estimation  by  learning  a
  regressor   from   an  input   image   to}   \bt{either  directly   the   pose
  vectors~\cite{Toshev14}    or   the}    \ms{heatmaps    encoding   2D    joint
  locations~\cite{Jain14,Pfister15,Tompson14}.    Recently,   this   trend   has
  extended to  3D pose estimation~\cite{Li14a},  where the problem  is typically
  formulated in terms  of continuous 3D joint locations,  since discretizing the
  3D space is more challenging than in the case of 2D.

Another important  difference between 2D and  3D pose estimation comes  from the
additional ambiguities  in the latter  one due to the  fact that the  input only
shows  a  projection  of  the  output. To  overcome  these  ambiguities,  recent
algorithms  have attempted  to  encode the  dependencies  between the  different
joints within  Deep Learning  approaches, thus effectively  achieving structured
prediction.  In particular,  \cite{Hong15} uses auto-encoders to  learn a shared
representation for 2D silhouettes and  3D poses.  This approach, however, relies
on accurate foreground  masks and exploits handcrafted  features, which mitigate
the  benefits  of Deep  Learning.   In  the  context  of hand  pose  estimation,
\cite{Oberweger15} introduces  a bottleneck, low-dimensional layer  that aims at
accounting for joint dependencies. This layer, however, is obtained directly via
PCA, which limits the kind of dependencies it can model.

To the  best of  our knowledge,  the work  of~\cite{Li15a} constitutes  the most
effective approach to encoding dependencies within a Deep Learning framework for
3D human pose estimation. This approach  extends the structured SVM model to the
Deep Learning setting by learning  a similarity score between feature embeddings
of the  input image  and the  3D pose. This  process, however,  comes at  a high
computational cost  at test  time, since,  given an  input image,  the algorithm
needs to  search for the  highest-scoring pose.  Furthermore, the  final results
are  obtained  by averaging  over  multiple  high-scoring ground-truth  training
poses, which might  not generalize well to unseen data  since the prediction can
thus  only be  in  the convex  hull  of  the ground-truth  training poses.   By
contrast,      we       draw      inspiration      from       the      KDE-based
approaches~\cite{Ionescu11,Ionescu14a},  that  map both  image  and  3D pose  to
high-dimensional Hilbert spaces and learn  a mapping between these spaces. Here,
however,  we show  how to  do this  in  a Deep  Learning context  with CNNs  and
auto-encoders.  The benefits are twofold: We  can leverage the power of learned
features  that  have  proven  more  effective  than  handcrafted  ones  such  as
HOG~\cite{Agarwal04a},  and  our framework  relies  on  a direct  and  efficient
regression between the two spaces, thus avoiding the computational burden of the
state-of-the-art approach of~\cite{Li15a}.

Using auto-encoders  for unsupervised feature  learning has proven  effective in
several  recognition  tasks~\cite{Konda15,Kingma14,Vincent10}.   In  particular,
denoising auto-encoders~\cite{Vincent08} that aim  at reconstructing the perfect
data  from a  corrupted  version  of it  have  demonstrated good  generalization
ability.  Similarly,  contractive  auto-encoders  have  been  shown  to  produce
intermediate representations  that are robust  to small variations of  the input
data~\cite{Rifai11}.  All these methods, however, rely on auto-encoders to learn
features for recognition  tasks. By contrast, here, we  exploit auto-encoders to
model the output structure for regression purposes.  }

\section{Method}
\label{sec:method}

In this  work, we aim  at directly regressing  from an input  image $x$ to  a 3D
human pose.  As in~\cite{Bo10,Ionescu14,Li14a}, we represent the human pose in terms
of the  3D locations $y  \in \mathbb{R}^{3J}$ of $J$  body joints relative  to a
root joint.   An alternative would  have been to  predict the joint angles  and limb
lengths, however this is a less  homogeneous representation and is therefore rarely used
for regression.

%

\begin{figure}[t]
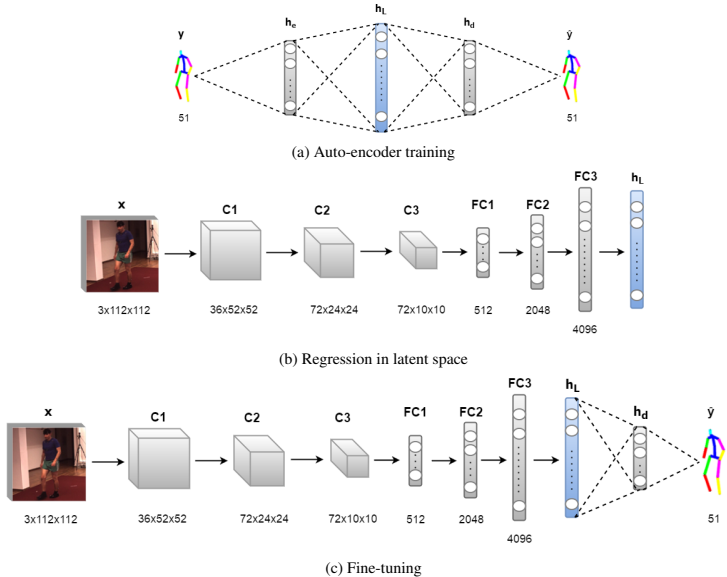

	\centering
	\scalebox{0.75}{
		\begin{tabular}{c}
			\includegraphics[width=0.6\linewidth]{aetraining} \\
			\footnotesize (a) Auto-encoder training \\
			\includegraphics[width=0.8\linewidth]{CNNembedding} \\
			\footnotesize (b) Regression in latent space \\   
			\includegraphics[width=\linewidth]{CNNpose}   \\
			\footnotesize (c) Fine-tuning \\
		\end{tabular}
	}
	\caption[Full training  of the  CNN]{Our approach.  {\bf(a)} We  train a
          stacked  denoising  auto-encoder  that learns  and  enforces  implicit
          constraints about human body in  its latent middle layer $h_{L}$. {\bf
            (b)}  Our   CNN  architecture   maps  the   image  to   the  latent
          representation $h_{L}$ learned by the auto-encoder. {\bf (c)} We stack
          the  decoding  layers of  the  auto-encoder  on  top  of the  CNN  for
          reprojection  from the  latent space  to the  original pose  space and
          fine-tune  the  entire  network  by updating  the  parameters  of  all
          layers.}
	\vspace{-2mm}
	\label{fig:full_network}
\end{figure}

As discussed  above, a straightforward  approach to  creating a regressor  is to
train a  conventional CNN such as  the one used in~\cite{Li14a}.   However, this
fails  to encode  dependencies between  joint locations.   In~\cite{Li15a}, this
limitation  was  overcome by  introducing  a  substantially more  complex,  deep
architecture  for  maximum-margin  structured  learning.   Here,  \ms{we  encode
  dependencies in a simpler, more efficient, and ultimately more accurate way by
  learning  a mapping  between the  output of  a conventional  CNN and  a latent
  representation obtained using an  overcomplete auto-encoder, as illustrated in
  Fig.~\ref{fig:full_network}.  The  auto-encoder is pre-trained on  human poses
  and comprises  a hidden  layer of  {\it higher dimension}  than its  input and
  output.  In effect, this hidden layer  and the CNN-based representation of the
  image   play  the   same  role   as   the  kernel   embeddings  in   KDE-based
  approaches~\cite{Cortes05,Ionescu11,Ionescu14a}, thus  allowing us  to account
  for structure within  a direct regression framework. Once  the mapping between
  these two  high-dimensional embeddings  is learned,  we further  fine-tune the
  whole network for the final pose estimation task, as depicted at the bottom of
  Fig.~\ref{fig:full_network}.}

%
%
%

\ms{In the remainder of this section, we describe the different stages of our approach.}

\vspace{-2mm}

\subsection{Using Auto-Encoders to Learn Structured Latent Representations}
\label{sec:encoder}


We encode  the dependencies  between human  joints by learning  a mapping  of 3D
human pose to a  high-dimensional latent space. To this end,  we use a denoising
auto-encoder that can have one or more hidden layers. 

Following  standard  practice~\cite{Vincent10}, given  a  training  set of  pose
vectors  $\{y_i\}$, we  add isotropic  Gaussian noise  to create  noisy versions
$\{\tilde{y}_i\}$ of these  vectors.  We then train our auto-encoder  to take as
input  a noisy  $\tilde{y}_i$  and  return a  denoised  $y_i$ as  output. The
  corresponding reconstruction function $f_{ae}(\cdot)$ must satisfy
\begin{equation}
\hat{y} = f_{ae}(y,\theta_{ae}) \;, 
\label{eq:reconstruct}
\end{equation}
\noindent where $\hat{y}$ is the reconstruction and $\theta_{ae} =  (W_{enc,j}, b_{enc,j}, W_{dec,j},  b_{dec,j})_{j=1}^L$ contains
the model parameters, that is, the  weights and biases for $L$ encoding and decoding
layers. We  take the  middle layer to be our latent pose  representation and denote 
it by $h_L$. We use ReLU as the activation function of the encoding layer. 
This  favors a  sparse hidden representation~\cite{Glorot11a},  
which has been   shown   to  be   effective   at   modeling   a   wide  range   of   
human poses~\cite{Akhter15,Ramakrishna12}. A linear activation function is used at the 
decoding layer of the auto-encoder to reproject to both negative and positive joint
coordinates. To keep the number of parameters small and reduce overfitting,  we 
use tied weights  for the encoder and  the decoder, that is, $W_{dec,j}=W_{enc,j}^T$.  
\comment{, which can be expressed as
\begin{equation}
h_L = g(W_{enc} y + b_{enc})\;.
\label{eq:hidden}
\end{equation} }

\pf{To learn the parameters $\theta_{ae}$, we rely on the square loss between the reconstruction, $\hat{y}$, and the 
original input, $y$, over the $N$ training examples. To increase robustness to small pose
changes, we regularize the cost function by adding the squared Frobenius norm 
of the  Jacobian of the hidden mapping $g(\cdot)$, that is, $J(y)=\frac{\partial g}{\partial y}(y)$ where $g(\cdot)$
is the encoding function that maps the input $\tilde{y}$ to the middle hidden layer, $h_L$. Training can thus 
be expressed as finding 
\begin{equation}
\theta_{ae}^* = \argmin_{\theta_{ae}} \sum_i^N ||y_i-f(y_i,\theta_{ae})||_2^2 + \lambda \| J(y_i) \|_F^2 \;,
\label{eq:aecost}
\end{equation}
where $\lambda$ is the regularization weight.}
Unlike when using KDE, we  do not need to solve a complex
pre-image problem to go from the  latent pose representation to the pose itself.
\ms{This mapping,  which corresponds to  the decoding part  of our  auto-encoder, is
   learned directly from data.}

\subsection{Regression in Latent Space}

\ms{Once the auto-encoder is trained, we aim to learn a mapping between the input image and the latent representation of the human pose. To this end, and as shown in Fig.~\ref{fig:full_network}(b), we make use of a CNN to regress the image to a high-dimensional representation, which is itself mapped to the latent pose representation.

More specifically, let $\theta_{cnn}$ be the parameters of the CNN, including the mapping to the latent pose representation. Given an input image $x$, we consider the square loss function between the representation predicted by the CNN, $f_{cnn}(x,\theta_{cnn})$, and the one that was previously learned by the auto-encoder, $h_L$. Given our $N$ training samples, learning amounts to finding}
\begin{equation}
\theta_{cnn}^* = \argmin_{\theta_{cnn}} \sum_i^N ||f_{cnn}(x_i,\theta_{cnn})-h_{L,i}||_2^2 \;.
\label{eq:regress}
\end{equation}

\ms{
In practice, as shown in Fig.~\ref{fig:full_network}(b), we rely on a  standard   
CNN  architecture  similar  to  the  one of~\cite{Li14a,Toshev14}.  It  comprises three
convolutional layers---C1,  C2 and C3---each  followed by a  pooling layer---P1,
P2, and P3.  In our implementation, the input volume is a three channel image of
size  $128\times128$. $P3$  is directly  connected to  a cascade  of fully-connected
layers---FC1,  FC2 and  FC3---that produces a 4096-dimensional image representation,
 which is then mapped linearly to the latent pose embedding. Except for this last 
linear layer, each  layer uses a ReLU activation function.

As in~\cite{Li14a}, prior to training our CNN, we first initialize the convolutional layers
 using a network trained for the detection of body joints in 2D.} \ms{We then replace the 
 fully-connected layers of the detection network with those of the regressor
 to further train for the pose estimation task.}

\vspace{-2mm}

\subsection{Fine-Tuning the Whole Network}
\label{sec:ft}

\ms{
Finally, as shown in Fig.~\ref{fig:full_network}(c), we append the decoding layers of the auto-encoder to the CNN discussed above, which reprojects the latent pose estimates to the original pose space.
We then fine-tune the resulting complete network for the task of human pose estimation. We take the cost function to be the squared difference between the predicted and ground-truth 3D poses, which yields the optimization problem
\begin{equation}
\theta_{ft}^* = \argmin_{\theta_{ft}} \sum_i^N ||f_{ft}(x_i,\theta_{ft})-y_i||_2^2 \;,
\label{eq:regress}
\end{equation}
\noindent where $\theta_{ft}$ are the complete set of model parameters and $f_{ft}$ is the
 mapping function.}

\vspace{-5mm}

\section{Results}

In this section, we first describe the large-scale dataset we tested our approach on. We 
then give implementation details and describe the evaluation protocol. Finally, we
compare our results against those of the state-of-the-art methods.

\vspace{-2mm}

\subsection{Dataset}

We evaluate our  method on the Human3.6m dataset~\cite{Ionescu14a}, 
which comprises 3.6 million  image frames with their corresponding  2D and 3D
poses. The subjects perform complex motion scenarios based on typical 
human activities such as  discussion,  eating, greeting and walking. 
The videos were captured from 4 different camera viewpoints. Following the standard procedure of~\cite{Li14a}, we collect the
input images by extracting  a square region around the subject using  
the bounding box present in the dataset and resize  it to $128 \times 128$. 
The  output pose is  a vector of $17$ 3D joint coordinates. 

\vspace{-2mm}

\subsection{Implementation Details}

We trained our auto-encoder using a greedy layer-wise training scheme followed by
fine-tuning   as   in~\cite{Hinton06b,Vincent10}. We set the regularization weight of Eq.~\ref{eq:aecost} to
$\lambda =  0.1$. We experimented with single-layer auto-encoders, as well as with $2$-layer ones. 
The size of the layers were set to 2000 and 300-300 for the 1-layer and 2-layer cases, respectively. 
We  corrupted the  input pose  with zero-mean Gaussian noise with standard deviation of 40 for 
1-layer and 40-20 for 2-layer auto-encoders. In all cases, we used the ADAM optimization 
procedure~\cite{Kingma15} with a learning rate of $0.001$ and a batch 
size of $128$.

The number and individual sizes of the layers of our CNNs are given in
	Fig.~\ref{fig:full_network}.
The  
	filter sizes for the convolutional layers are consecutively  $9
	\times 9$, $5  \times 5$ and $5 \times 5$.  Each convolutional layer
	is followed  by a  $2 \times  2$ max-pooling layer.  The activation
	function is  the ReLU in all  the layers except  for the last one that uses
	linear activation. As for the auto-encoders, we  used ADAM~\cite{Kingma15}  with a learning
	rate  of $0.001$  and a  batch size of $128$.  To prevent
	overfitting, we applied dropout with a probability of $0.5$ after each
	fully-connected layer and augment the data by randomly cropping $112
	\times 112$ patches from the $128 \times 128$ input images.

\subsection{Evaluation Protocol}

For a fair comparison, we used the same data partition protocol as in earlier 
work~\cite{Li14a,Li15a} for training and test splits. The data from $5$ subjects (S1,S5,S6,S7,S8) 
was used for training and the data from $2$ different subjects (S9,S11) was used for testing.
We evaluate the accuracy of 3D human pose estimation in terms of average Euclidean
distance between the predicted and ground-truth 3D joint positions as 
in~\cite{Li14a,Li15a}. The accuracy numbers are reported in milimeters for all actions
on which the authors of~\cite{Li14a,Li15a} provided results. Training and testing 
were carried out monocularly in all camera views for each separate action.

\subsection{Experimental Results}

\begin{figure}[t]
	\centering
	\scalebox{1}{
		\begin{tabular}{c}
			\includegraphics[width=2.1in, height=0.75in]{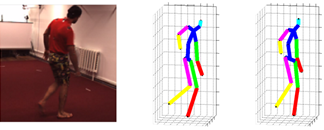} \hspace{1cm}			 
			\includegraphics[width=2.1in, height=0.75in]{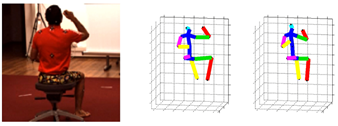} \\
			\includegraphics[width=2.1in, height=0.75in]{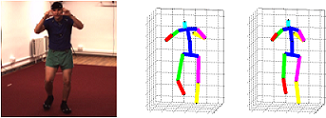} \hspace{1cm}
			\includegraphics[width=2.1in, height=0.75in]{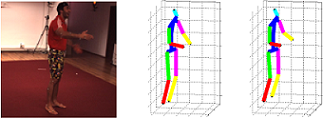} \\
			\includegraphics[width=2.1in, height=0.75in]{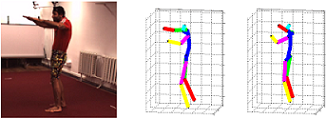} \hspace{1cm}
			\includegraphics[width=2.1in, height=0.75in]{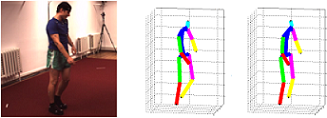} \\
		\end{tabular}
	}
	\caption{3D  poses  for  the   \textit{Walking,  Eating,  Taking  Photo,
            Greeting,  Discussion}  and  \textit{Walking  Dog}  actions  of  the
          Human3.6m  database.   In  each   case,  the  first  skeleton  depicts
          the ground-truth  pose and  the second  one the pose  we recover.   Best viewed  in
          color.}
	\vspace{-2mm}
	\label{fig:prediction_vis}
\end{figure}

\begin{table}[t]
	\centering
	\scalebox{0.7}{
		\begin{tabular}{lcccccc}
			\toprule
			Model											&Discussion	&Eating	&Greeting	&Taking Photo	&Walking	&Walking Dog  \\ 
			\midrule
			LinKDE(~\cite{Ionescu14a}						&183.09		&132.50	&162.27	&206.45			&97.07		&177.84	   \\
			DconvMP-HML~\cite{Li14a}							&148.79		&104.01 &127.17	&189.08			&77.60		&146.59    	\\
			StructNet-Max~\cite{Li15a}							&149.09 	&109.93 &136.90	&179.92			&83.64		&147.24	    \\
			StructNet-Avg~\cite{Li15a}				&134.13 	&97.37 	&122.33	&166.15			&68.51		&132.51    \\
			\emph{OURS}							&{\bf 129.06}		&\bf{91.43}	&\bf{121.68}		&\bf{162.17}		&\bf{65.75}		&\bf{130.53}  \\	
			\bottomrule 
		\end{tabular}
	}
	\caption{Average  Euclidean distance  in  mm  between the ground-truth 3D joint
          locations       and      those       predicted      by       competing
          methods~\cite{Ionescu14a,Li14a,Li15a} and ours.}
    \vspace{-3mm}
	\label{tab:sota}
\end{table}

\setlength{\tabcolsep}{2pt}
\begin{table}[t]
	\centering
	\scalebox{0.68}{
		\begin{tabular}{lcccccc}
			\toprule
			Model											&Discussion	&Eating	&Greeting	&Taking Photo	&Walking	&Walking Dog  \\ 
			\midrule
			\emph{CNN-Direct}													&135.36 	&105.98	&133.35	&177.62			&77.73 		&153.02  	\\	
			\emph{OURS, 1 layer no FT}					 					&134.02 	&96.01	&127.58	&158.73			&68.55		&146.28    \\
			\emph{OURS, 2 layer no FT}									&129.67 	&98.57	&124.80		&162.69			&73.47		&146.46   \\
			\emph{OURS, 1 layer with FT}						&130.07 &94.08 &121.96 &\bf{158.51}	&{65.83}	&135.35  \\
			\emph{OURS, 2 layer with FT}								&\bf{129.06}		&\bf{91.43}	&\bf{121.68}		&162.17		&\bf{65.75}		&\bf{130.53}  \\	
			\bottomrule 
		\end{tabular}
	} 
	\scalebox{0.68}{
		\begin{tabular}{lc}
			\toprule
			Model 	 & \pf{Taking Photo} \\   
			\midrule
			\emph{CNN-Direct}						 &177.62\\ 
			\emph{CNN-ExtraFC[2000]}		 &179.29\\
			\emph{CNN-PCA[30]}			     &170.74\\ 
			\emph{CNN-PCA[40]}			     &167.62\\ 
			\emph{CNN-PCA[51]}			     &182.64\\ 
			\emph{OURS[40]}					 &165.11\\ 
			\emph{OURS[2000]}			     &\bf{158.51}\\ 
			\bottomrule 
		\end{tabular}
	} \\
	\footnotesize \hspace{2.7cm} {\bf(a)} \hspace{5.4cm}  {\bf(b)}
        \vspace{-0.3cm}
        \caption{Average Euclidean  distance in  mm between the  ground-truth 3D
          joint locations and  those computed {\bf (a)} using  either no auto-encoder
          at all (CNN)  or 1-layer and 2-layer encoders (OURS),  with or without
          fine-tuning (FT), {\bf (b)} by replacing the auto-encoder by either
            an additional  fully-connected layer  (\emph{CNN-ExtraFC}) or  a PCA
            layer (\emph{CNN-PCA}).   The bracketed  numbers denote  the various
            dimensions of  the additional layer  we tested.  Our  approach again
            yields the most accurate predictions.}
        \vspace{-5mm}
	\label{tab:autoenc}
\end{table}
\setlength{\tabcolsep}{6pt}


Fig.~\ref{fig:prediction_vis}  depicts  selected   pose  estimation  results  on
Human3.6m.  In Table~\ref{tab:sota}, we report our results on this dataset along
with  those  of  three  state-of-the-art approaches:  KDE  regression  from  HOG
features to  3D poses~\cite{Ionescu14a}, jointly  training a body  part detector
and  a  3D  pose  regressor network~\cite{Li14a},  and  using  a  maximum-margin
formalism    within     a    Deep    Learning    framework     for    structured
prediction~\cite{Li15a}. For  the latter, the  estimation is taken to  be either
the  highest-scoring pose  or the  average of  the 500  highest-scoring training
poses. Our method consistently outperforms all the baselines.

The results  reported in  Table~\ref{tab:sota} were obtained  using a  two layer
auto-encoder. However,  as discussed in Section~\ref{sec:encoder}  our formalism
applies  to auto-encoders  of any  depth.  Therefore,  in Table~\ref{tab:autoenc}
(a), we report results  obtained using a single layer one as  well as by turning
off the  final fine-tuning of  Section~\ref{sec:ft}.  For completeness,  we also
report   results   obtained  by   using   a   CNN   similar   to  the   one   of
Fig.~\ref{fig:full_network}(b) to  regress directly to a  51-dimensional 3D pose
vector {\it without} using an auto-encoder at all. We will refer to it as \emph{
  CNN-Direct}.  We found that both kinds of auto-encoders perform similarly and
better than  \emph{ CNN-Direct},  especially for  actions such  as \emph{Taking
  Photo} and \emph{Walking  Dog} that involve interactions  with the environment
and are thus  physically more constrained.  This confirms that  the power of our
method  comes   from  auto-encoding.   Furthermore,  as   expected,  fine-tuning
consistently improves the results.

During  fine-tuning, our complete network  has more fully-connected layers  than 
  \emph{ CNN-Direct}. One could therefore argue that  the additional layers are
  the reason  why our approach outperforms  it.  To disprove this,  we evaluated
  the  baseline,  \emph{CNN-ExtraFC}, in which we simply add one more fully-connected
  layer. We also evaluated another baseline, \emph{CNN-PCA}, in which we  replace  
  our  auto-encoder  latent  representation  by  a
  PCA-based  one.   In Table~\ref{tab:autoenc}(b),  we  show  that our  approach
  significantly outperforms  these two baselines  on the {\it  Taking Photo} action.
  This suggests that our overcomplete  auto-encoder yields a representation that
  is  more discriminative  than  other  latent ones.   Among  the different  PCA
  configurations,  the  one with  40  dimensions  performs the  best.   However,
  training an auto-encoder with 40 dimensions outperforms it.

\begin{figure}[t] 
	\centering 
	\scalebox{0.8}{
		\begin{tabular}{cc}
			\includegraphics[width=0.45\columnwidth]{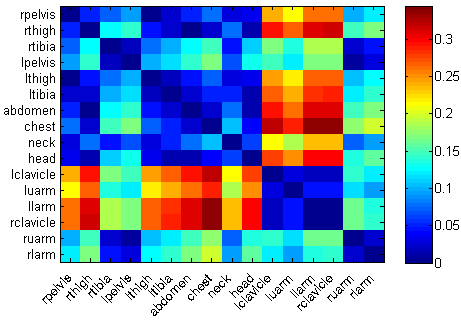} 
			\includegraphics[width=0.45\columnwidth]{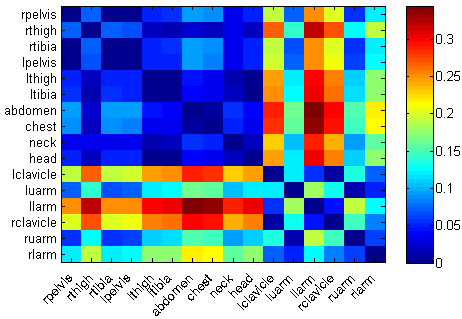}\\
			\footnotesize (a) \hspace{5.5cm} \footnotesize (b) \\
			\includegraphics[width=0.45\columnwidth]{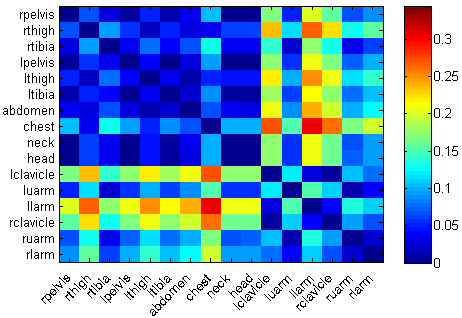} 
			\includegraphics[width=0.45\columnwidth]{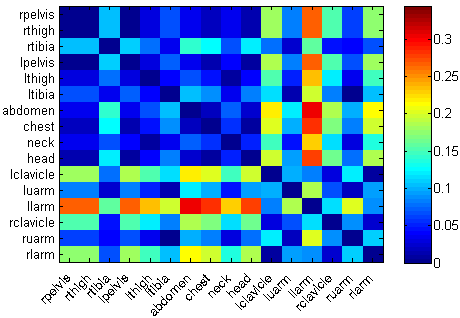}\\
			\footnotesize (c) \hspace{5.5cm}	\footnotesize (d) \\
		\end{tabular}
	} \\
	\scalebox{0.7}{
		\begin{tabular}{l c c c}
			\toprule 
			Model 	&Lower Body &Upper Body & Full Body \\ 
			\midrule
			\emph{KDE~\cite{Ionescu14a}}		&1.02 &7.18 &16.43 \\
			\emph{CNN}					&0.57 &6.86 &14.97	\\
			\emph{OURS no FT}						&0.62 &5.30 &11.99\\
			\emph{OURS with FT}				&0.77 &5.43 &11.90\\
			\bottomrule 
		\end{tabular}
	} \\
	\footnotesize (e)
	\vspace{-4mm}
	\caption{Matrix of differences between  estimated log of limb length ratios
          and  those computed  from ground-truth  poses.  The  rows and  columns
          correspond  to  individual  limbs.   For each  cell,  the  ratios  are
          computed by dividing the limb length in the horizontal axis by the one
          in   the  vertical   axis   as  in~\cite{Ionescu11}   for  {\bf   (a)}
          KDE~\cite{Ionescu14a},  {\bf (b)}  CNN-Direct as  in Table~\ref{tab:autoenc},
          and {\bf  (c,d)} our method  without and  with fine-tuning.   An ideal
          result would  be one  in which  all cells are  blue, meaning  the limb
          length ratios  are perfectly  preserved.  Best  viewed in  color. {\bf
            (e)} Sum of the log of limb  length ratio errors for different parts of
          the  human   body.   All  methods   perform  well  on   the  lower
            body. However,  ours outperforms  the others on  the upper  body and
            when considering all ratios in the full body.}
	\label{fig:LLLR}
	\vspace{-4mm}
\end{figure}

Following~\cite{Ionescu11}, we  show in Fig.~\ref{fig:LLLR}  the differences
  between  the  ground-truth limb  ratios  and  the  limb ratios  obtained  from
  predictions based  on KDE, \emph{CNN-Direct}  and our approach.  These results
  evidence  that our  predictions better  preserve these  limb ratios,  and thus
  better model the dependencies between joints.

\vspace{-3mm}

\subsection{Parameter Choices}

In  Table~\ref{tab:component_eval},   we  compare   the  results   of  different
auto-encoder configurations in terms of number  of layers and channels per layer
on    the    {\it   Greeting}    action.     Similarly    to   what   we did    in
Table~\ref{tab:autoenc}(b), the  bracketed numbers  denote the dimension  of the
auto-encoder's hidden layers.  We obtained the  best result for 1 layer with 2000
channels or  2 layers  with 300-300 channels.  These values  are those we used for 
all  the experiments described above. They were  chosen for a  single action  and 
used  unchanged for all  others, thus demonstrating the versatility of our approach.


\begin{table}[t]
	\label{tab:ae_exp}
	\centering
	\scalebox{0.8}{
		\begin{tabular}{l c}
			\toprule
			Layer Configuration 	& Greeting \\ 
			\midrule
			\emph{[40]}						&129.49\\
			\emph{[500]}					&123.95\\
			\emph{[1000]}					&121.96\\
			\emph{[2000]}					&\bf{121.96}\\
			\emph{[3000]}					&123.49\\
			\emph{[250-250]}				&125.61\\
			\emph{[300-300]}	 			&\bf{121.68}\\
			\emph{[250-500]}				&128.98\\
			\emph{[500-1000]}	 			&126.52\\
			\emph{[200-200-200]}	 		&126.78\\
			\emph{[500-500-500]}	 		&127.73\\
			\bottomrule 
		\end{tabular}
	} 
	\\
    	\caption{Average Euclidean  distance in  mm between the  ground-truth 3D
          joint locations and  the ones predicted by our  approach trained using
          auto-encoders  in various  configurations,  with  different number  of
          layers and number of channels per  layer as indicated by the bracketed
          numbers. This  validation was performed  on the {\it  Greeting} action
          and the optimal values used for all other actions.}
	\label{tab:component_eval}
\end{table}


\vspace{-5mm}

\section{Conclusion}

We have introduced a novel  Deep Learning regression architecture for structured
prediction of  3D human  pose from  a monocular  image. We  have shown  that our
approach  to combining  auto-encoders  with CNNs  accounts  for the  dependencies
between the human  body parts efficiently and yields  better prediction accuracy
than state-of-the-art methods.  Since our  framework is generic, in future work,
we intend to apply it to other structured prediction problems, such as deformable 
surface reconstruction.

\clearpage
{\small
	\bibliographystyle{bmvc2k}
	\bibliography{string,short,vision,learning}
}

\end{document}